\documentclass[runningheads]{llncs}

\usepackage{graphicx}
\usepackage{amsmath}
\usepackage[T1]{fontenc}
\usepackage{hyperref}
\usepackage{footnote}
\usepackage{booktabs}
\usepackage{bm}
\usepackage{multirow}
\usepackage[utf8]{inputenc}
\makesavenoteenv{tabular}
\makesavenoteenv{table}

\newcommand{\bx}{\mathbf{x}}
\newcommand{\bxp}{\mathbf{x'}}
\newcommand{\bxt}{\mathbf{\tilde{x}}}

\begin{document}
\title{NeuralHydrology - Interpreting LSTMs in Hydrology}

\titlerunning{Interpreting LSTMs in Hydrology}

\author{Frederik Kratzert\inst{1}\thanks{1st corresponding author: kratzert@ml.jku.at} \and
Mathew Herrnegger\inst{2}\thanks{2nd corresponding author: mathew.herrnegger@boku.ac.at} \and
Daniel Klotz\inst{1} \and Sepp Hochreiter\inst{1} \and G\"unter Klambauer\inst{1}}

\authorrunning{F. Kratzert et al.}

\institute{LIT AI Lab \& Institute for Machine Learning\\ Johannes Kepler University Linz \\ A-4040 Linz, Austria \bigskip \and
Institute of Hydrology and Watermanagement \\ University of Natural Resources and Life Sciences, Vienna\\ A-1190 Vienna, Austria}

\maketitle

\begin{abstract}
Despite the huge success of Long Short-Term Memory networks,
their applications in environmental sciences are scarce.
We argue that one
reason is the difficulty to interpret the internals of trained networks.
In this study, we look at the application of LSTMs for rainfall-runoff forecasting, one of the central tasks in the field of hydrology, in which the
river discharge has to be predicted from meteorological observations.
LSTMs are particularly well-suited for this problem since memory
cells can represent dynamic reservoirs and storages, which are essential components in state-space modelling approaches of the hydrological system.
On basis of two different catchments, one with snow influence and one without, we demonstrate how the trained model can be analyzed and interpreted.
In the process, we show that the network internally learns to represent patterns
that are consistent with our qualitative understanding of the hydrological system.

\end{abstract}

\setcounter{footnote}{0}
\renewcommand{\thefootnote}{\roman{footnote}}

\section{Introduction}
Describing the relationship between rainfall and runoff is one of the central tasks in the field of hydrology \cite{Klemes1982}.
This involves the prediction of the river discharge from meteorological observations from a river basin.
The basin or catchment of a river is defined by the area of which all (surface) runoff drains to a common outlet \cite{WMO2012}.
Predicting the discharge of the river is necessary for e.g. flood forecasting, the design of flood protection measures, or the efficient management of hydropower plants.

\begin{figure}
  \centering
  \includegraphics[width=11cm]{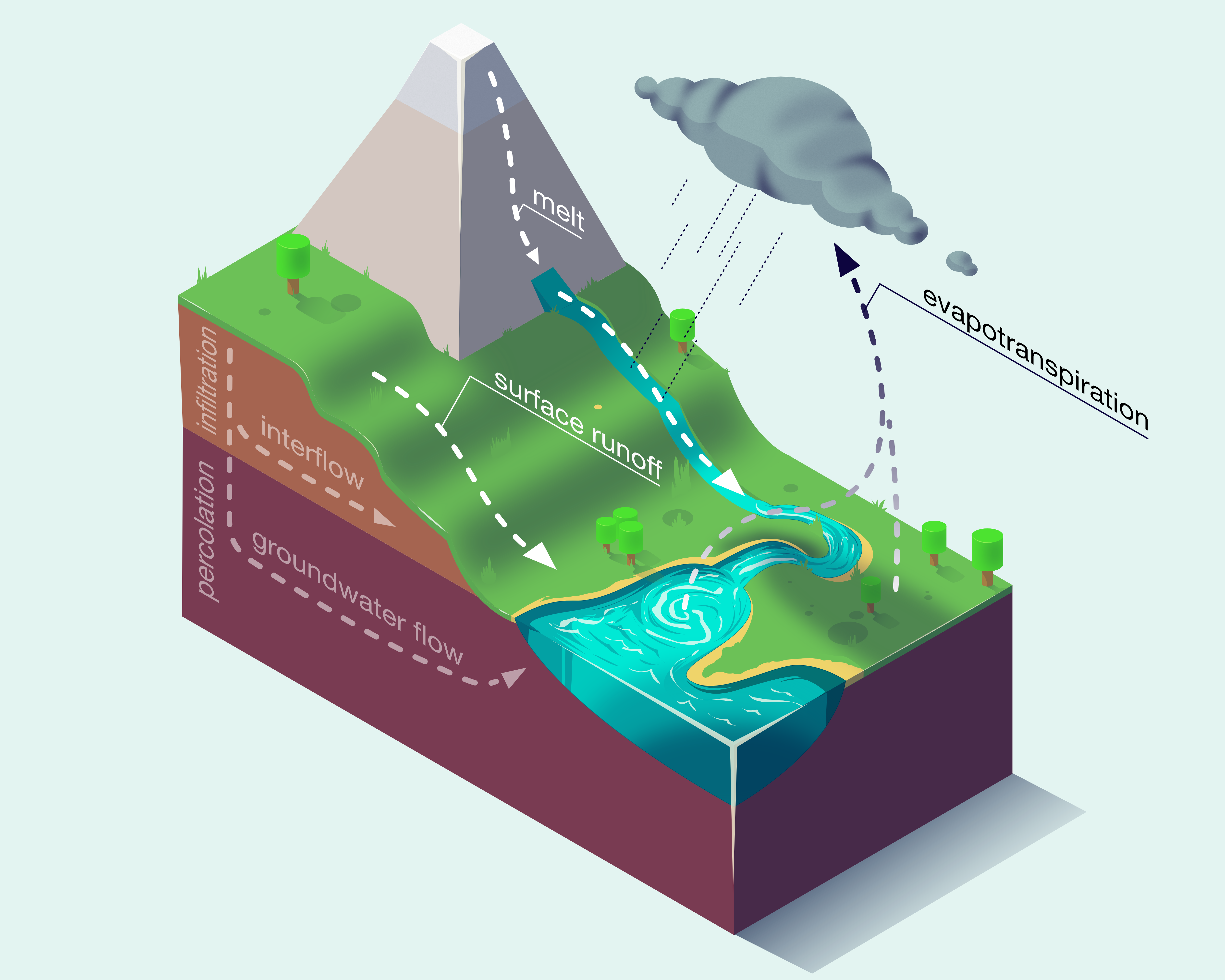}
  \caption{Simplified visualization of processes and fluxes that influence the river
  discharge, such as precipitation, snow melt, surface runoff or subsurface flows.} \label{fig_hyd_mod}
\end{figure}

Within the basin of a river, various hydrological processes take place that influence and lead to the river discharge,
including, for example, evapotranspiration, where water is lost to the atmosphere, snow accumulation and
snow melt, water movement in the soil or groundwater recharge and discharge (see Fig. \ref{fig_hyd_mod}).

The hydrological processes have highly non-linear interactions and depend, to a large degree, on the states of
the system, which represent the memory of, e.g. a river basin. Consequently, hydrological models
are formulated in a state-space approach where the states at a specific time depend on the
input $\boldsymbol{I}_t$, the system states at the previous time step $\boldsymbol{S}_{t-1}$, and a
set of parameters $\Theta_i$ \cite{Herrnegger2015}:

\begin{equation}\label{eq_1}
  \boldsymbol{S}_t = f(\boldsymbol{I}_t, \boldsymbol{S}_{t-1}; \Theta_i)
\end{equation}

The discharge at a given time step $t$ is driven by the system states and in consequences by the meteorological events of the preceding time steps. More generally, any output $\boldsymbol{O}_t$ of a hydrological system (e.g. the runoff) can be described as:

\begin{equation}\label{eq_2}
  \boldsymbol{O}_t = g(\boldsymbol{I}_t, \boldsymbol{S}_{t}; \Theta_j),
\end{equation}

where $g(\cdot)$ is a mapping function that connects the states of the system and the inputs to the system output,
and $\Theta_j$ is the corresponding subset of model parameters.

For making proficient predictions these non-linearities make it inevitable (at least in classical process-based hydrological models)
to explicitly implement the hydrological processes \cite{Herrnegger2012,Lindstrom2010,Perrin2003,Thielen2008}.
However, defining the mathematical representations of the processes, including the model structures and
determining their effective parameters so that the resulting system exhibits good performance and
generalizable properties (e.g. in the form of seamless parameter fields) still remains a
challenge in the field of hydrology \cite{Gupta1999,Klotz2017,Samaniego2017}.

A significant problem and limiting factor in this context is the frequently missing information regarding
the physical properties of the system \cite{Beven2001,Freeze1969}. These tend to be highly heterogeneous
in space (e.g. soil characteristics) and can additionally change over time, e.g. vegetation cover.
Our knowledge of the properties on, or near the surface has increased significantly in the last decades.
This is mainly due to advances in high-resolution air- as well as spaceborne remote
sensing \cite{Brenner2017,Hengl2017,Myneni2002}. However, hydrology, to a significant part,
takes place underground, for which detailed information is rarely available.
In essence, the process-based models try to describe a system determined by spatially and
temporally distributed system states and physical parameters, which are most of the time unknown.

In contrast, data-driven methods, such as Neural Networks, are solely trained to predict the
discharge, given meteorological observations, and do not necessitate an explicit definition
of the underlying processes. But these models have not the best reputation among many
hydrologists because of the prevailing opinion that models ``must work well for the right reasons'' \cite{Klemes1986}.
However, due to their predictive power, first studies of using Neural Networks
for predicting the river discharge date back to the early 90s \cite{Daniell1991,Halff1993}.

Recently, Kratzert et al. \cite{hess-2018-247} used Long Short-Term Memory networks (LSTMs) \cite{Hochreiter1997}
for daily rainfall-runoff modelling and could show that LSTMs achieve competitive results, compared to the well established Sacramento Soil Moisture Accounting Model \cite{Burnash1973} coupled with the Snow-17 snow model \cite{Anderson1973}.
LSTM is an especially well-suited network architecture for Hydrological applications, since the evolution of states can be modelled explicitly through time and mapped to a given output.
The approach is very similar to rainfall-runoff models defined by
Equations \ref{eq_1}-\ref{eq_2} (in the case of the LSTM the system states are
the memory cell states and the parameters are the learnable network weights, \cite{hess-2018-247}).

The aim of this chapter is to show different possibilities that enable the interpretation
of the LSTM and its internals in the context of rainfall-runoff simulations. Concretely,
we explore and investigate the following questions: How many days of the past  influence
the output of the network at a given day of the year? Do some of the memory cells correlate
with hydrological states? If yes, which input variables influence these cells and
how? Answering these questions is important to a) gain confidence in
data-driven models, e.g. in case of the necessity for extrapolation, b) have tools
to understand possible mistakes and difficulties in the learning process
and c) potentially learn from findings for future applications.

\subsection{Related work}
In the field of water resources and hydrology, a lot of effort has been made on interpreting neural networks and analyzing the importance of input variables (see \cite{Bowden2005} for an overview).
However, so far only feed-forward neural networks have been applied in these studies. Only recently, Kratzert et al. \cite{hess-2018-247} have demonstrated the potential use of LSTMs for the task of rainfall-runoff modelling. In their work they have also shown that memory cells with interpretable functions exist, which were found by visual inspection.

Outside of hydrology, LSTMs found a wide range of applications, which attracted researches to study on analyzing and interpreting the network internals. For example Hochreiter et al. \cite{Hochreiter2007} found new protein motifs through analyzing LSTM memory cells.
Karpathy et al. \cite{Karpathy2015} inspected memory cells in character level language modelling and identify some interpretable memory cells, e.g. cells that track the line-length or cells that check if the current text is inside brackets of quotation marks.
Li et al. \cite{Li2015} inspected trained LSTMs in the application of sentence- and phrase-based sentiment classification and showed through saliency analysis, which parts of the inputs are influencing the network prediction most.
Arras et al. \cite{Arras2017} used Layer-wise Relevance Propagation to calculate the impact of single words on sentiment analysis from text sequences.
Also for sentiment analysis, Murdoch et al. \cite{Murdoch2018} present a decomposition strategy for the hidden and cell state of the LSTM to extract the contributions of single words on the overall prediction.
Poerner et al. \cite{Poerner2018} summarize various interpretability efforts in the natural language processing domain and present an extension of the LIME framework, introduced originally by Reibiero et al. \cite{Ribeiro2016}.
Strobelt et al. \cite{Strobelt2018} developed LSTMVis, a general purpose visual analysis tool for inspecting hidden state values in recurrent neural networks.

Inspired by these studies, we investigate the internals of LSTMs in the domain of environmental science and compare our findings to hydrological domain knowledge.

\section{Methods}
\subsection{Model architecture}
In this study, we will use a network consisting of a single LSTM layer with 10 hidden
units and a dense layer, that connects the output of the LSTM at the last time step
to a single output neuron with linear activation. To predict the discharge of a single time step (day) we provide the last 365 time steps of meteorological observations as inputs.
Compared to Eq. \ref{eq_1} we can formulate the LSTM as:

\begin{equation}\label{eq_3}
  \{\boldsymbol{c}_t, \boldsymbol{h}_t\} = f_{\mathrm{LSTM}}(\boldsymbol{i}_t, \boldsymbol{c}_{t-1}, \boldsymbol{h}_{t-1}; \Theta_k),
\end{equation}

where $f_{\mathrm{LSTM}}(\cdot)$ symbolizes the LSTM cell that is a function of the meteorological
input $\boldsymbol{i}_t$ at time $t$, and the previous cell state $\boldsymbol{c}_{t-1}$ as
well as the previous hidden state $\boldsymbol{h}_{t-1}$, parametrized by the network weights $\Theta_k$.
The output of the system, formally described in Eq. \ref{eq_2}, would in this specific case be given by:

\begin{equation}\label{eq_4}
  y = f_{\mathrm{Dense}}(\boldsymbol{h}_{365}; \Theta_l),
\end{equation}

where $y$ is the output of a dense layer $f_{\mathrm{Dense}}(\cdot)$ parametrized by the
weights $\Theta_l$, which predicts the river discharge from the hidden state at the end
of the input sequence $\boldsymbol{h}_{365}$.

The difference between the LSTM and conventional rainfall-runoff models is that the
former has the ability to infer the needed structure/parametrization from data without
preconceived assumptions about the nature of the processes. This makes them
extremely attractive for hydrological applications.

The network is trained for 50 epochs to minimize the mean squared error using
RMSprop \cite{Tieleman2012} with an initial learning rate of 1e-2. The final model is
selected based on the score of an independent validation set.

\subsection{Data}

In this work, we concentrate on two different basins from the
publicly available CAMELS data set \cite{Addor2017,Newman2014}. Basin A,
which is influenced by snow, and basin B, which is not influenced by snow.
Some key attributes of both basins can be found in Table~\ref{tab_basin}.

\begin{table}
\caption{Basin overview.}\label{tab_basin}
\setlength{\tabcolsep}{10pt}
\begin{tabular}{ c  c c  c  c  c }
\multirow{2}{*}{Basin} & \multirow{2}{*}{ID\footnote{USGS stream gauge ID}} & Snow & \multirow{2}{*}{Area (km\textsuperscript{2})} & NSE & NSE \\
 &  & fraction\footnote{Fraction of precipitation falling with temperatures below 0$^{\circ}$C} &  & validation & test\\

\toprule
A & 13340600\footnote{Clearwater river, CA} & 56 \% & 3357 & 0.79  & 0.76 \\
B & 11481200\footnote{Little river, CA} & 0 \% & 105 & 0.72  & 0.72 \\
\end{tabular}
\end{table}

For meteorological forcings, the data set contains basin averaged daily records
of precipitation (mm/d), solar radiation (W/m\textsuperscript{2}), minimum and
maximum temperature ($^{\circ}$C) and vapor pressure (Pa). The streamflow is
reported as daily average (m\textsuperscript{3}/s) and is normalized by the
basin area (m\textsuperscript{2}) to (mm/d). Approximately 33~years of data
is available, of which we use the first 15 for training the LSTMs. Of the
remaining years the first 25~\% is used as validation data by which we select
the final model. The remaining data points (approx. 13 years) are used for the
final evaluation and for all experiments in this study. The meteorological input
features, as well as the target variable, the discharge are normalized by the
mean and standard deviation of the training period.

One LSTM is trained for each basin separately and the trained model is evaluated
using the Nash-Sutcliffe-Efficiency \cite{Nash1970}, an established measure used to
evaluate hydrological time series given by the following equation:

\begin{equation}\label{eq_5}
  \text{NSE} = 1 - \frac{\sum_{t=1}^{T}(Q_m^t - Q_o^t)^2}{\sum_{t=1}^{T}(Q_o^t - \bar{Q}_{o})^2},
\end{equation}

where $T$ is the total number of time steps, $Q_m^t$ is the simulated discharge at
time $t$ $(1 \leq t \leq T)$, $ Q_o^t$ is the observed discharge at time $t$ and $\bar{Q}_{o}$ is
the mean observed discharge. The range of the NSE is (-inf, 1], where a value of 1 means a
perfect simulation, a NSE of 0 means the simulation is as good as the mean of the
observation and everything below zero means the simulation is worse compared to
using the observed mean as a prediction.

In the test period the LSTM achieves a NSE of above 0.7 (see Table \ref{tab_basin}),
which can be considered a reasonably good result \cite{Moriasi2015}.

\begin{figure}
\includegraphics[width=\textwidth]{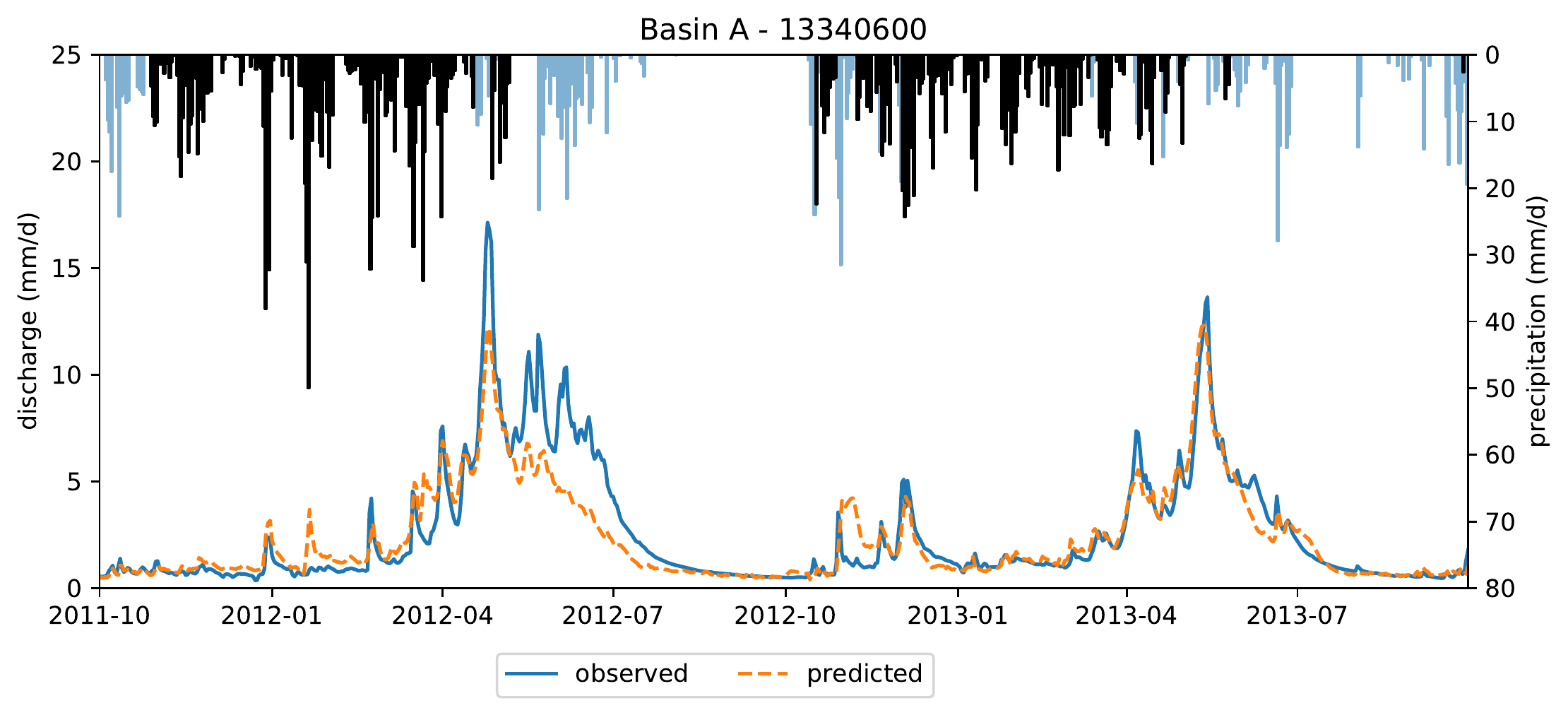}
\caption{Example of predicted (dashed line) and observed discharge (solid line) of
two years of the test period in the snow influenced basin. Corresponding daily
precipitation sums are plotted upside down, where snow
(temperature below 0$^{\circ}$C) is plotted darker. \label{fig_hydrograph}}
\end{figure}

Figure \ref{fig_hydrograph} shows observed and simulated discharge of two years of
the test period in the snow influenced basin A, as well as the input variable precipitation.
We can see that the discharge has its peak in the spring/early summer und that the model in the
year 2012 underestimates the discharge, while in the second year it fits the observed
discharge pretty well. The time lag between precipitation and discharge can be
explained by snow accumulation in the winter months and subsequent melt of this snow layer in spring.

\subsection{Integrated gradients}
Different methods have been presented recently to analyze the attribution of input variables on the network output (or any in-between neuron) (e.g. \cite{Arras2017,Bach2015,Kindermans2017,Montavon2017,Sundararajan2017})

In this study we focus on Integrated gradients by Sundarajan et al. \cite{Sundararajan2017}.
Here, the attribution of each input to e.g. the output neuron is calculated
by looking at the change of this neuron when the input shifts from a baseline
input to the target input of interest. Formally, let $\boldsymbol{x}$
be the input of interest, (in our case a sequence of 365 time steps with 5
meteorological observations each), ${\boldsymbol{x}}'$ the baseline input
and $F(\cdot)$ the neuron of interest. Then the integrated gradients, for the \textit{i}-th input variable $x_i$, can be appoximated by:

\begin{equation}\label{eq_6}
  \text{IntegratedGrads}_i^{\mathrm{approx}}(\bm x) := \frac{\bx_i - \bxp_i}{m} \sum_{k=1}^{m}\frac{\partial F\left(\bxt \right)}{\partial \bxt_i}\bigg|_{\bxt = \bxp + \frac{k}{m}(\bx - \bxp)},
\end{equation}

where $m$ is the number of steps used to approximate the integral (here $m=1000$).
As baseline ${\boldsymbol{x}}'$, we used an input sequence of zeros.

\subsection{Experiments}

\subsubsection{Question 1: How many days of past influence the network output?\newline}
The discharge of a river in a seasonal influenced region varies strongly throughout the year.
For e.g. snow influenced basins the discharge usually peaks in the spring or early summer,
when not only precipitation and groundwater but also snow melt contributes to the discharge generation.
Therefore, at least from a hydrological point of view, the precipitation of the entire
winter might be influential for the correct prediction of the discharge. In contrast,
in drier periods (e.g. here at the end of summer) the discharge likely depends on far
fewer time steps of the meteorological past. Since we provide a constant number of
time steps (365 days) of meteorological data as input, it is interesting to see how
many of the previous days are really used by the LSTM for the final prediction.

To answer this question, we calculate the integrated gradients for one sample
of the test period w.r.t. the input variables and sum the integrated gradients across
the features for each time step. We then calculate the difference from time
step to time step and determine the first timestep $t$ $(1 \leq t \leq T)$,
at which the difference surpasses a threshold of 2e-3, with $T$ being the
total length of the sequence. We have chosen the threshold value empirically
so that noise in the integrated gradient signal is ignored. For each sample the
number of Time Steps Of Influence (TSOI) on the final prediction can then be calculated by:

\begin{equation}\label{eq_7}
  \text{TSOI} = T - n
\end{equation}

This is repeated for each day of each year in the test period.

\subsubsection{Question 2: Do memory cells correlate with hydrological states?\newline}
The discharge of a river basin is frequently approximated by decomposing its
(hypothetical) components into a set of interacting reservoirs or storages
(see Fig. \ref{fig_hyd_mod}). Take snow as an example, which is precipitation that falls if temperatures are below 0$^{\circ}$C. It  can be represented in a
storage $\boldsymbol{S}$ (see Eq.~\ref{eq_1}), which generally accumulates
during the winter period and depletes from spring to summer when the temperatures
rise above the melting point. Similarly other components of the system can
be modelled as reservoirs of lower or higher complexity. The soil layer,
for example, can also be represented by a bucket, which is filled -
up to a certain point - by incoming water (e.g. rainfall) and depleted
by evapotranspiration, horizontal subsurface flow and water movement
into deeper layers, e.g. the groundwater body.

Theoretically, memory cells of LSTMs could learn to mimic these storage
processes. This is a crucial property, at least from a hydrological point of view, to be able to correctly predict the river discharge. Therefore the aim of this experiment
is to see if certain memory cells $\boldsymbol{c}_t$ (Eq.~\ref{eq_3})
correlate to these hydrological states $\boldsymbol{S}_t$  (Eq.~\ref{eq_1}).

Because the CAMELS data does not include observations for these states,
we take the system states of the included SAC-SMA + Snow-17 model as a proxy.
This is far from optimal, but since this is a well established and studied
hydrological model, we can assume that at least the trend and tendencies of
these simulations are correct. Furthermore, we only want to test in this experiment if memory cells correlate with these system states and not if they
quantitatively match these states exactly. Of the calibrated SAC-SMA + Snow-17
we use the following states as a reference in this experiment:

\begin{itemize}
\item \textbf{SWE} (snow water equivalent): This is the amount of water stored
in the snow layer. This would be available if the entire snow in the system would melt.

\item \textbf{UZS} (upper zone storage): This state is calculated as the sum of
the UZTWC (upper zone tension water storage content) and the UZFWC (upper zone
free water storage) of the SAC-SMA + Snow-17 model. This storage represents upper
layer soil moisture and controls the fast response of the soil, e.g. direct surface runoff and interflow.

\item \textbf{LZS} (lower zone storage):  This state is calculated by the sum of
LZTWC (lower zone tension water content), LZFSC (lower zone free supplemental water
storage content) and LZFPC (lower zone free primary water storage content).
This storage represents the groundwater storage and is relevant for the
baseflow\footnote{Following \cite{WMO2012} the baseflow is defined as:
``Discharge which enters a stream channel mainly from groundwater, but also
from lakes and glaciers, during long periods when no precipitation or snowmelt occurs.''}.
\end{itemize}

For each sample in the test period we calculate the correlation of the cell states with the corresponding time series of these four states.

\subsubsection{Question 3: Which inputs influence a specific memory cell?\newline}

Suppose that we find memory cells that correlate with time series of hydrological system states,
then a natural question would be if the inputs influencing these memory cells agree with
our understanding of the hydrological system. For example, a storage that represents the snow layer
in a catchment, should be influenced by precipitation and solar radiation in a contrarious
way. Solid precipitation or snow would increase the amount of snow available in the system
during winter. At the same time, solar radiation, providing energy for sublimation, would
effectively reduce the amount of snow stored in the snow layer.
Therefore, in this experiment we look in more detail at specific cells that emerged from
the previous experiment and analyse the influencing variables on this cell. We do this by
calculating the integrated gradients from a single memory cell at the last time step of the input sequence w.r.t. the input variables.

\section{Results and Discussion}

\subsection{Timesteps influencing the network output}

\begin{figure}
  \centering
  \includegraphics[width=12cm]{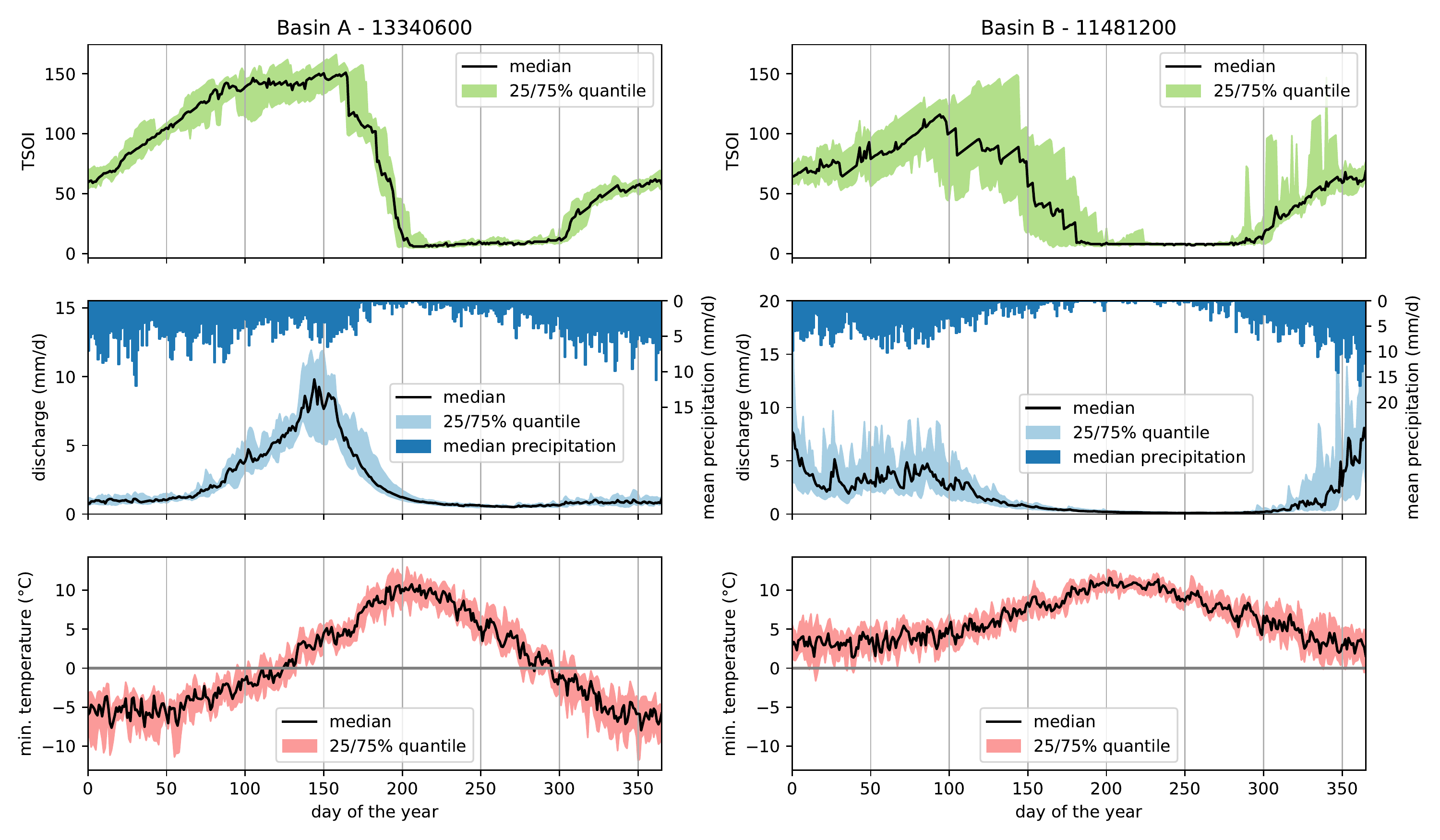}
  \caption{Time steps of influence (TSOI) on the network output over the unit day of year (DOY) for the snow influenced basin A (left column) and the basin B without snow influence (right column). Corresponding median precipitation, discharge and minimum temperature are shown for reference.
  For the snow-influenced basin A, we can see for example that the TSOI increases during the winter period and is largest during the snow melting period ($\sim$DOY 100-160), which matches our understanding of the hydrological processes.
  } \label{fig_tsoi}
\end{figure}

Figure \ref{fig_tsoi} shows how many time steps with meteorological inputs from the past have an influence on the LSTM output at the time step of prediction (TSOI). The TSOI does
thereby not differentiate between single inputs. It is rather the integrated signal of all inputs.
Instead of using specific dates, we here show the temporal dimension in the unit day of
year (DOY). Because all years of the test period are integrated in the plot, we show the 25~\%, 50~\% and 75~\% percent
quantiles. For the sake of interpretation and the seasonal context, Fig. \ref{fig_tsoi} also includes the temporal dynamics of the median precipitation, temperature and discharge.

The left column of Fig. \ref{fig_tsoi} shows the results for snow influenced basin A. Here, 3 different periods can be distinguished in the TSOI time series:

(1) Between DOY 200 and 300 the TSOI shows very low values of less than 10-20~days. This period corresponds
to the summer season characterised by high temperatures and low flows, with fairly little precipitation.
In this period the time span of influence of the inputs on the output is short. From a hydrological
perspective this makes sense, since the discharge in this period is driven by short-term rainfall events,
which lead to a very limited response in the discharge. The short time span of influence can be explained by higher evapotranspiration
rates in this season and lower precipitation amounts. Higher evapotranspiration rates lead to the loss of precipitation to the atmosphere, which is then missing in the discharge.
The behaviour of the hydrograph is fairly easy to predict and does not necessitate much information about the past.

(2) In the winter period, starting with DOY 300, the TSOI increases over time reaching a plateau of 140-150 days around DOY 100. In this period the daily minimum
temperature is below 0$^{\circ}$C, leading to solid precipitation (snow) and therefore water
being stored as snow in the catchment without leading to runoff. This is underlined with the
low discharge values, despite high precipitation input. Thus the LSTM has to understand
that this input does not lead to an immediate output and therefore the TSOI has to increase.
The plateau is reached, as soon as the minimum temperature values are higher than the freezing point.
From a hydrological perspective, it is interesting to observe that the TSOI at the end of the
winter season has a value which corresponds to the beginning of the winter season ($\sim$DOY 300),
when the snow accumulation begins. It should be noted that the transition between the winter
period and the following spring season is not sharp, at least when taking the hydrograph as
a reference. It is visible that, although the TSOI is still increasing, the discharge is also
increasing. From a hydrological point of view this can be explained by a mixed signal in
discharge - appart from melting snow (daily maximum temperature is larger than 0$^{\circ}$C),
we still have negative minimum temperatures, which would lead to snow fall.

(3) In spring, during the melting season, the TSOI stays constant (DOY 100-160) followed
by a sharp decrease until the summer low flows. During the main melting period, the TSOI of approximately
140-150 days highlights that the LSTM uses the entire information of the winter period to
predict the discharge. The precipitation in this period now falls as rain, directly
leading to runoff, without increasing the TSOI. At the same time, all the inputs
from the winter still influence the river discharge explaining the stable plateau.
The sharp decrease of the TSOI around DOY 160 represents a transition
period where the snow of the winter continuously loses its influence until all snow has melted.

Although it has the same precipitation seasonality, Basin B (Fig. \ref{fig_tsoi}, right column) has different characteristics compared
to basin A, since it is not influenced by snow.
Here, only 2 different periods can be distinguished, where the transitions periods are however more pronounced:

(1) Between DOY 180 and 280, the warm and dry summer season, the catchment is
characterised by very low flows. In this period the TSOI is also constantly low,
with values of around 10-15 days. The discharge can be predicted with a very short
input time series, since rainfall as input is missing and the hydrology does not depend on any inputs.

(2) Following summer, the wet period between DOY 280 and 180 is characterised by
a steady increase in the TSOI. The temporal influence of rainfall on runoff becomes
longer, the longer the season lasts. The general level of runoff is now significantly
higher compared to the summer. It does not solely depend on single rainfall events,
but is driven by the integrated signal of inputs from the past, explaining the
increasing TSOI. The TSOI reaches a maximum median value of around 120 days. This
peak is followed by a decrease towards the end of the wet period, which is however
not as rapid as in basin A. This distinct transition period in TSOI between wet
and dry ($\sim$DOY 100-180) season corresponds very well with the observed falling
limb in the hydrograph. As the runoff declines, the influence of past meteorological inputs also declines.
Compared to basin A, a higher variability in TSOI is evident, which can be explained
with a higher variability in precipitation inputs in the single years. In basin A,
a high interannual variability in the rainfall is also observable. However, the
lower temperatures below freezing level lead to the precipitation falling as snow,
and therefore act as a filter. This leads to a lower variability in discharge and in consequence in TSOI.

Overall, the TSOI results of the two contrasting basins match well with our hydrological
understanding of the anterior days influencing the runoff signal at a specific day.
It is interesting to see that the LSTM shows the capability to learn these differing, basin specific properties of long-term dependencies.

\subsection{Correlation of memory cells with hydrological states}

\begin{figure}
  \centering
  \includegraphics[width=\textwidth]{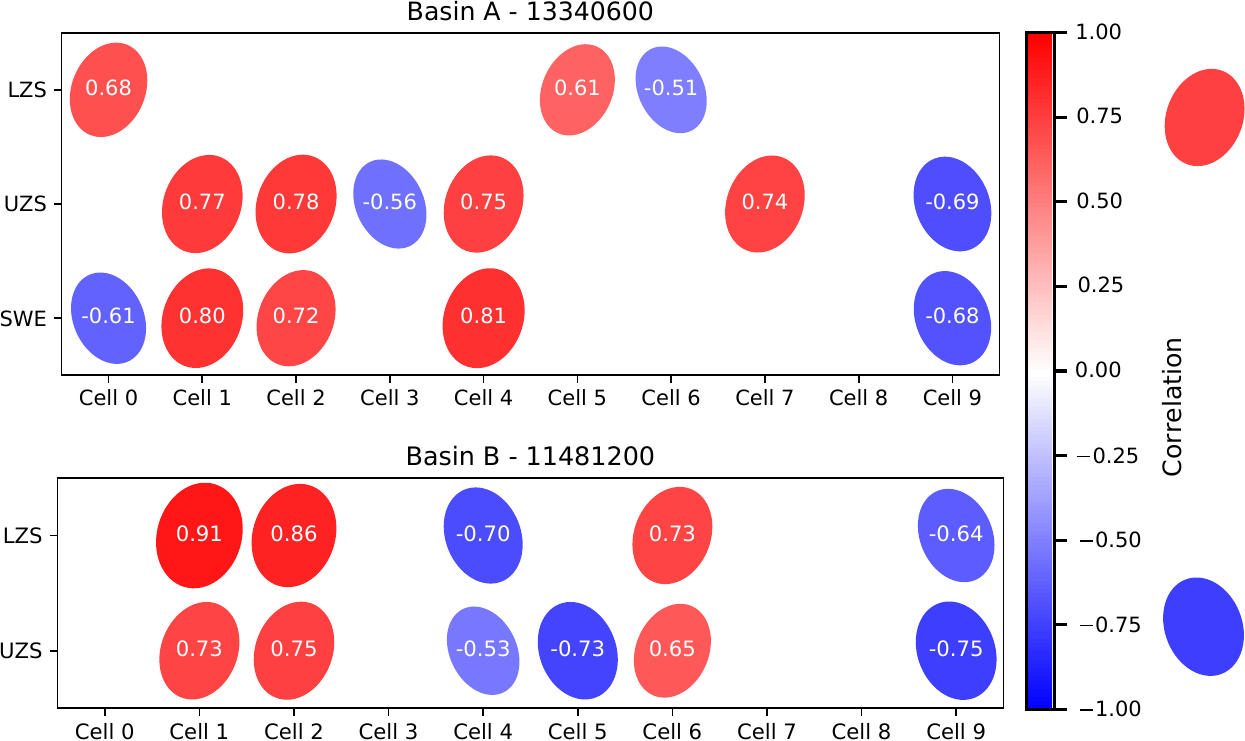}
  \caption{Average correlations between memory cells and hydrological states for
  basin A (upper plot) and basin B (lower plot). Only correlations with $ \rho > 0.5$ are shown. Ellipse are scaled by the absolute correlation and
  rotated by their sign (left inclined for negative correlation and right inclined
  for positive correlation)} \label{fig_corr}
\end{figure}

Figure \ref{fig_corr} shows the average correlation of every memory cell with
the hydrological states considered in this experiment. The correlation is averaged
over all samples in the test period and only correlations with  $ \rho > 0.5$ are shown. We can see, that in both basins some of the memory cells have a particularly
high correlation with the provided hydrological states. For both basins several cells
exhibit a high correlation with both, the upper (UZS) and lower (LZS), soil states.
Although the majority of the cells show a positive correlation, negative correlations
are also visible, which are however of a lower absolute magnitude.

The correlation between the LSTM cells and the baseflow influencing state LZS are
significantly higher for the drier basin B. However, the baseflow index, a measure
to define the importance of the contribution of the baseflow to total runoff, is
lower for this basin. In basins A and B the ratios of mean daily baseflow to mean
daily discharge are about 66~\% and 44~\%, respectively. Currently, we cannot explain this discrepancy.
In the snow influenced basin, the trained LSTM also has some cells with
high correlation with the snow-water-equivalent (SWE). The occurrence of multiple
cells with high correlation to different system states can be seen as an indicator
that the LSTM is not yet defined in a parsimonious way. Therefore, hydrologist can use
this information to restrict the neural network even further.

In general, the correlation analysis is difficult to interpret in detail. Frequently, high correlations however exist, indicating a strong relationship between LSTM
memory cells and system states from a well-established hydrological model.

\subsection{Inspection of memory cells}

\begin{figure}
  \centering
  \includegraphics[width=10cm]{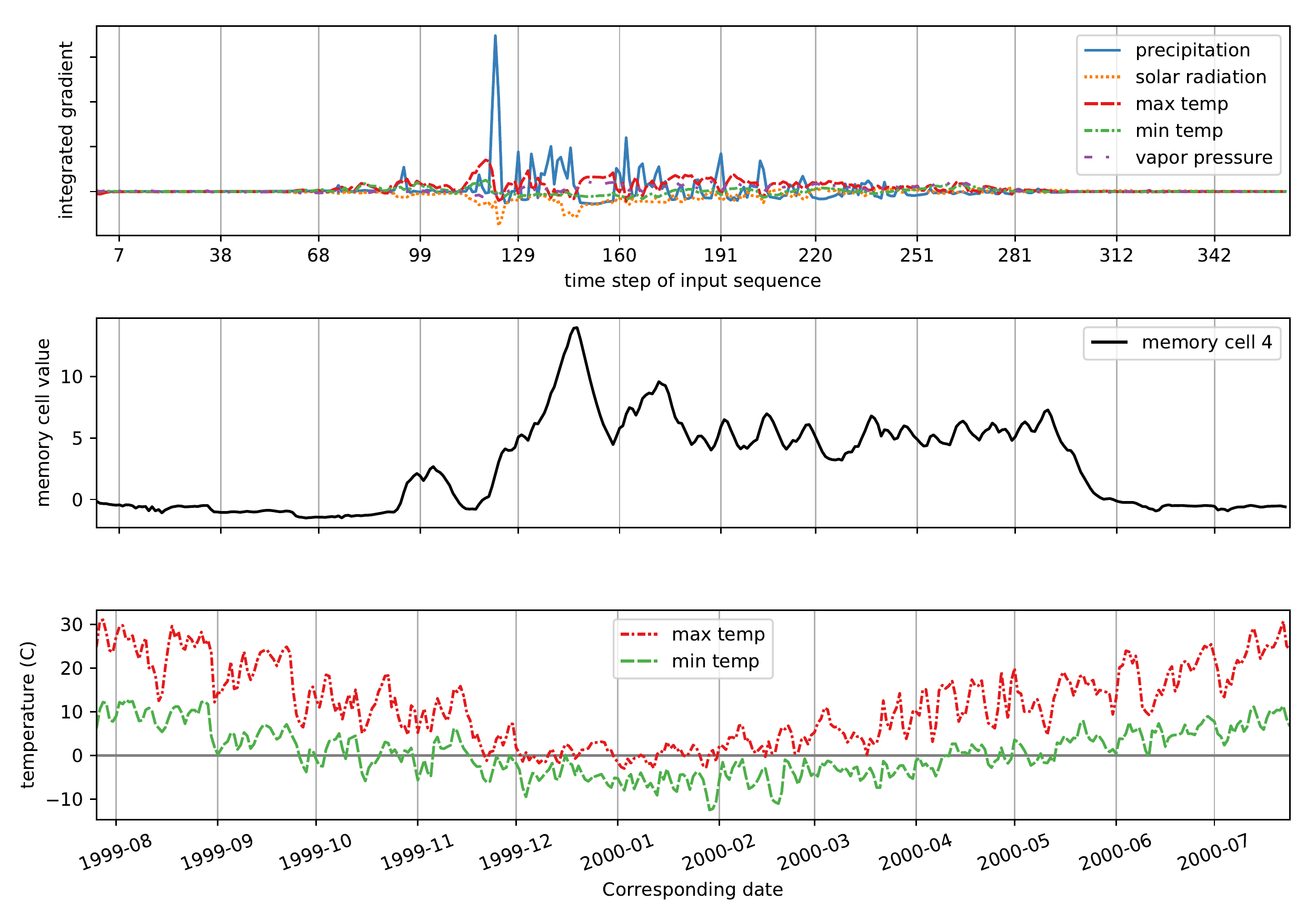}
  \caption{Integrated gradient analysis of the snow-cell (cell 4) of the LSTM
  trained for basin A. Upper plot shows the integrated gradient signal on each
  input variable at each time step. The plot in the center shows the memory cell
  state over time for reference and the bottom plot the minimum and maximum daily
  temperature.} \label{fig_snowcell}
\end{figure}

In the first experiment, we used the integrated gradient method to calculate the
attribution of the input variables on the output of the model. In this experiment,
we apply it to analyse interactions between arbitrary neurons within the neural
network (e.g. here, a memory cell at the last time step). The previous experiment
proved that memory cells with a high correlation to some of the
hydrological system states exist. The aim of this experiment is therefore to
analyse the influences and functionality of a given cell. Here, we can explore
(i) which (meteorological) inputs are important and (ii) at what time in the past
this influence was high. We chose a ``\textit{snow-cell}'' from the previous
experiment to demonstrate this task, since the accumulation and depletion of
snow is a particularly illustrative example. To be more precise, we chose to
depict a single sample of the test period from the cell with the highest
correlation to the SWE, which is memory cell 4 from basin A (see Fig. \ref{fig_corr}).
Figure \ref{fig_snowcell} shows the integrated gradients of the meteorological inputs in the top row, the
evolution of the memory cell value in the second row and the corresponding change in
minimum and maximum temperature in the third row.

One can see that the snow-cell is mainly influenced by the meteorological inputs
of precipitation, minimum temperature, maximum temperature and solar radiation.
Precipitation shows the largest magnitude of positive influence. Solar radiation
in contrast has a negative sign of influence, possibly reproducing the sublimation
from the snow layer and leading to a reduction of the snow state. All influencing
factors only play a role at temperatures around or below freezing level. This
matches the expected behaviour of a snow storage from a hydrological point of
view: Lower temperatures and concurrent precipitation are associated with
snow accumulation. Consequently, this leads to an increase in magnitude of the
memory cell value (especially for the first temperature below the freezing point).
This can be observed e.g. in October 1999, where the temperature values
decrease, the influences of the meteorological parameters appear and the snow-cell
begins to accumulate. In contrast, as the temperature rises, the value of the cell
decreases again, especially when the daily minimum temperature also rises above 0$^{\circ}$C.

This suggests that the LSTM realistically represents short- as well as long-term
dynamics in the snow cell storage and their connection to the meteorological inputs.

\section{Conclusion}

LSTM networks are a versatile tool for time series predictions,
with many potential applications in hydrology and environmental sciences in
general. However, currently they do not enjoy a wide-spread application. We argue
that one reason is the difficulty to interpret the LSTMs. The methods presented in
this book provide solutions regarding interpretability and allow a
deeper analysis of these models. In this chapter, we demonstrate this for the task
of rainfall-runoff modelling (where the aim is to predict the river discharge from
meteorological observations). In particular, we were able to show that the processes
learned by the LSTM matches our comprehension of a real-world environmental system.

For this study, we focused on a qualitative analysis of the correspondence between
the hydrological system and the learned behaviour of the LSTM. In a first experiment,
we looked at the number of time steps of influence on the network output (TSOI) and
how this number varies throughout the year. We saw that the TSOI pattern matches
our hydrological understanding of the yearly pattern. In the next experiment,
we looked at the correlation of the memory cells of the network with some selected
states of the hydrological system (such as snow or soil moisture). We found some
cells that exhibited relatively high correlation with the chosen states, which
strengthens the hypothesis that the LSTM obtained some general understanding of
the runoff-generation processes. In the last experiment, we inspected a single
memory cell that exhibited a high correlation with the snow state. We analyzed
the influencing inputs over time through the integrated gradient method and could
see, that the behavioral patterns manifested in the cell, closely resemble the
ones suggested by hydrological theory. We view this as a further underpinning
of the observation that the internally modelled processes of the network follow
some sort of physically viable pattern. We hypothesize that this relation can
be seen as a legitimization of the LSTM usage within environmental sciences
applications, and thus believe that the presented methods will pave the way
for its future in environmental-modelling. The correspondence of the memory cells
and the physical states can be especially useful in novel situations, which often
arise in this context. Environmental scientists and practitioners can exploit it
(together with the proposed techniques) to ``peek into the LSTM'' and argue about
potential behaviours. Our demonstration was certainly not exhaustive and
should rather be seen as indicative application study.

The most important message is that the combination of domain-knowledge (in this
case hydrology) and the insights provided by the proposed interpretation
techniques, provide the fundamentals for designing environmental forecasting
systems with neural networks. Consequently, we expect that the combination of
powerful data driven models (such as LSTMs) with the possibility of interpretation
by experts will lead to new insights in the field of application.

\bibliographystyle{splncs04}
\bibliography{bibliography}

\end{document}